\documentclass{article}

\usepackage{microtype}
\usepackage{graphicx}
\usepackage{subfigure}
\usepackage{amssymb}
\usepackage{amsmath}
\usepackage{paralist}
\usepackage{booktabs} 
\usepackage[margin=1cm]{caption}

\usepackage{hyperref}

\usepackage[accepted]{icml2018}

\icmltitlerunning{Attentive cross-modal paratope prediction}

\begin{document}

\twocolumn[
\icmltitle{Attentive cross-modal paratope prediction}



\icmlsetsymbol{equal}{*}

\begin{icmlauthorlist}
\icmlauthor{Andreea Deac}{cst}
\icmlauthor{Petar Veli\v{c}kovi\'{c}}{cst}
\icmlauthor{Pietro Sormanni}{chm}
\end{icmlauthorlist}

\icmlaffiliation{cst}{Department of Computer Science and Technology, University of Cambridge, UK}
\icmlaffiliation{chm}{Department of Chemistry, University of Cambridge, UK}

\icmlcorrespondingauthor{Andreea Deac}{aid25@cam.ac.uk}

\icmlkeywords{Machine Learning, ICML}

\vskip 0.3in
]



\printAffiliationsAndNotice{}  

\begin{abstract}
Antibodies are a critical part of the immune system, having the function of directly neutralising or tagging undesirable objects (the antigens) for future destruction. Being able to predict which amino acids belong to the \emph{paratope}, the region on the antibody which binds to the antigen, can facilitate antibody design and contribute to the development of personalised medicine. The suitability of deep neural networks has recently been confirmed for this task, with Parapred outperforming all prior models. Our contribution is twofold: first, we significantly outperform the computational efficiency of Parapred by leveraging \`{a} trous convolutions and self-attention. Secondly, we implement \emph{cross-modal attention} by allowing the antibody residues to attend over antigen residues. This leads to new state-of-the-art results on this task, along with insightful interpretations.
\end{abstract}

\section{Introduction}

Antibodies are Y-shaped proteins used by the immune system to neutralise pathogens such as bacteria and viruses. This is done when the antibody binds to the unique molecules on the pathogen called antigens. With antibodies being the most important class of biopharmaceuticals, knowing which amino acids are needed for the binding is a type of information that can have a significant impact on applications in diagnostics and therapeutics. In particular, creating novel antibodies requires the optimisation of properties such as solubility and stability, for which the non-binding amino acids can be used while maintaining the same binding affinity. 

Traditional attempts for predicting the binding amino acids (the \emph{paratope}) were based on hard coded physical models, requiring vast amounts of information. Predictors such as Antibody i-Patch \citep{i-patch} use as input the full structural information of the antibody and the antigen, while proABC \citep{proABC} requires the entire antibody sequence and additional features including the antigen volume. 

Only recently, Parapred \citep{Parapred}---a hybrid architecture consisting of convolutional and recurrent layers---has become the state of the art technique. However, its usage of recurrent layers represents a performance bottleneck, and it discards the information about the target antigen entirely.
 
In this work, we outperform Parapred by addressing its limitations and leveraging the bleeding-edge techniques in the language modelling community, such as \`{a} trous convolutions \citep{Atrous} and self-attention \citep{Self-attention}, while also significantly lowering computation time. We then manage to further improve this result by \emph{cross-modally attending} over sequential antigen information, managing to derive qualitative insights from the attentional coefficients in the process.

\section{Dataset and Preprocessing}
\label{data}


We used a subset of the Structural Antibody Database (SAbDab) \citep{Sabdab}, which provides crystal structures of antibody-antigen complexes, in order to train and evaluate our models. The subset was chosen under the same criteria as in \citet{Parapred}:
\begin{inparaenum}
	\item Antibodies having variable domains of their heavy ($V_H$) and light ($V_L$) chains;
	\item Structure resolution better than 3\r{A};
	\item No two antibody sequences have $>$95\% sequence identity;
	\item Each antibody has at least five amino acid residues\footnote{Throughout this document, we will frequently use the terms ``amino acid'' and ``residue'' interchangeably.} in contact with the target antigen.
\end{inparaenum} 

The paratope is contained within the \emph{complementarity determining regions} (\textbf{CDRs}) of the antibody. We identify the CDRs within the sequence of each antibody using the Chothia numbering scheme \cite{al1997standard}, and use each CDR as an independent training sequence.

For each residue in the CDR, we use the following features to obtain its feature vector, $\vec{ab}_i$:
\begin{itemize}
	\item A one-hot encoding of the \emph{amino acid type} (20 possible types + 1 additional for an unknown type);
	\item A one-hot encoding of the \emph{chain ID} of the CDR (6 possible types---three on the heavy chain (H1, H2, H3) and three on the light chain (L1, L2, L3). Consequentially, all residues within the same CDR will receive the same encoding;
	\item Seven additional features, summarised by \citet{meiler2001generation}, representing physical, chemical and structural properties of the given amino acid type (may be seen as a \emph{fixed embedding}).
\end{itemize}
In addition, for all of the complexes in the dataset, the antigens were \emph{proteins}. This allowed us to also extract the 1D residue sequences on the antigen. There is no equivalent for CDRs on the antigen, so the entire antigen sequence is extracted, and each residue's feature vector $\vec{ag}_i$ is obtained exactly as for $\vec{ab}_i$; omitting only the one-hot encoding of the chain ID, as the antigens aren't expected to have a fixed chain structure. The length of the longest CDR sequence is 32 residues, while for the longest antigen it is 1269 residues. Therefore, the shape of the largest possible CDR-antigen input to our model is \texttt{([32, 34], [1269, 28])}. 
\section{Methods}
\label{methods}

\subsection{Antibody-only}

We build up on the developments of Parapred by substituting its recurrent layers with a combination of \`{a} trous convolutional layers (for efficient modelling of longer-range dependencies) and a self-attentional layer (allowing for efficiently covering the sequence). We will refer to this architecture as \emph{Fast-Parapred} for the remainder of this paper.

\begin{figure}
	\includegraphics[width=\linewidth]{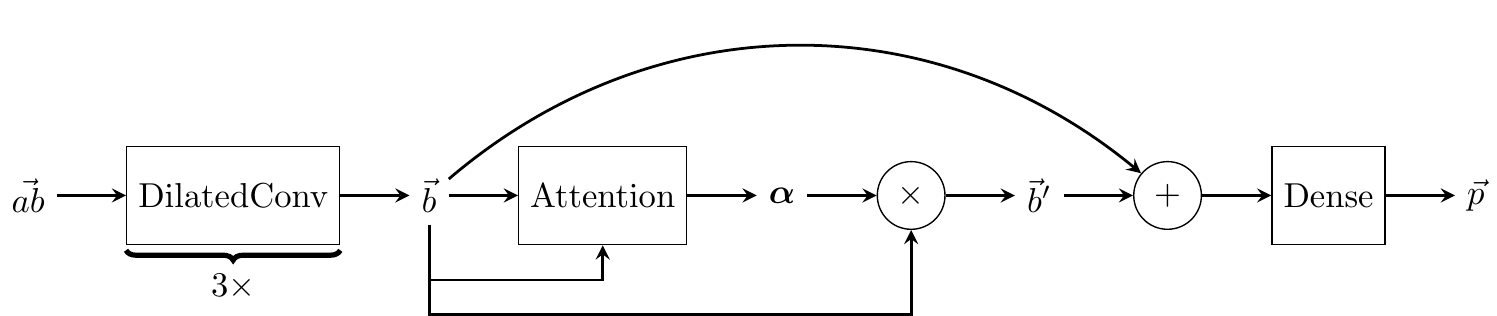}	
	\label{fastpar}
	\caption{The Fast-Parapred architecture.}
\end{figure}

A high-level layout of the architecture is presented in Figure 1. It receives as input the vector of antibody residue features $\vec{ab}$, and consists of:
\begin{itemize}
	\item A stack of three \`{a} trous (dilated) convolutional layers:
	\begin{enumerate}
		\item 64 features, kernel size 3, dilation rate 1;
		\item 128 features, kernel size 3, dilation rate 2;
		\item 256 features, kernel size 3, dilation rate 4.
	\end{enumerate}
	\item Then, a self-attention mechanism is applied on the computed intermediate features, $\vec{b}$.
	\item Lastly, a pointwise fully-connected (dense) layer is applied to classify each considered antibody amino acid as binding or non-binding.
\end{itemize}
All \`{a} trous convolutional layers and the self-attention layer employ the exponential linear unit (ELU) \citep{ELU} activation function, while the prediction layer uses the logistic sigmoid function to perform binary classification. All the layers are initialised using Xavier initialisation \cite{glorot2010understanding}.

\begin{figure}
	\centering
	\includegraphics[width=0.8\linewidth]{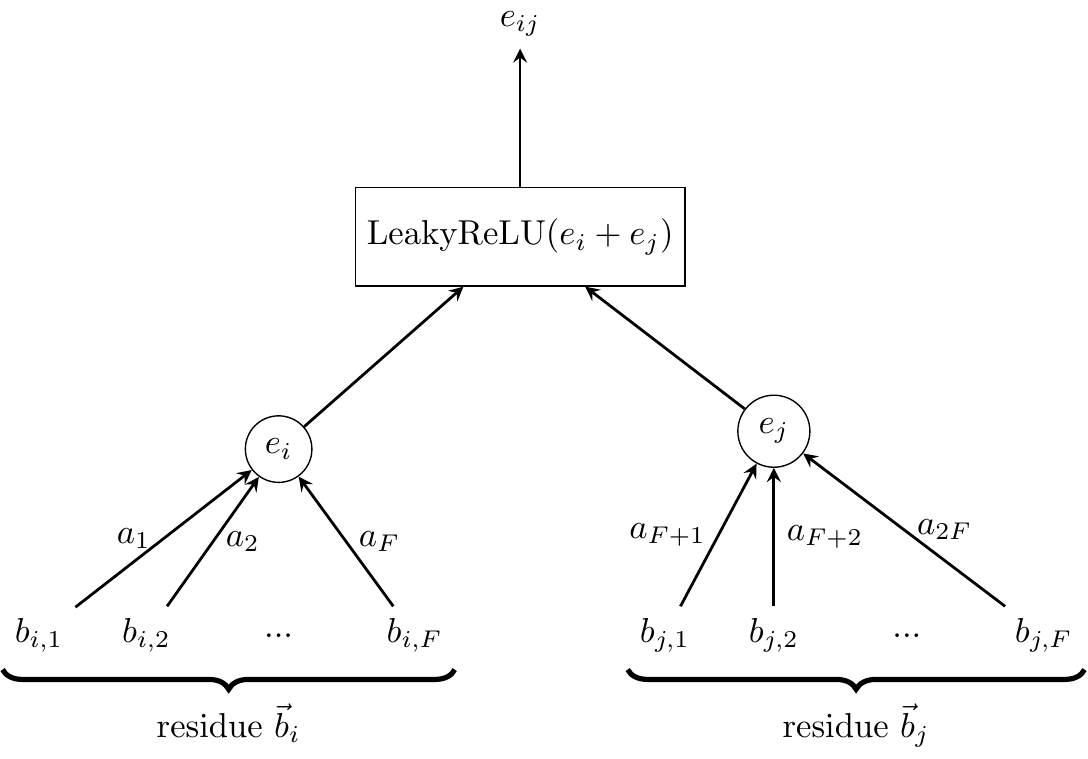}	
	\label{attnmech}
	\caption{The attentional mechanism $a$ within Fast-Parapred.}
\end{figure}

\begin{figure}
	\centering
	 \raisebox{9mm}[0pt][0pt]{\includegraphics[width=0.4\linewidth]{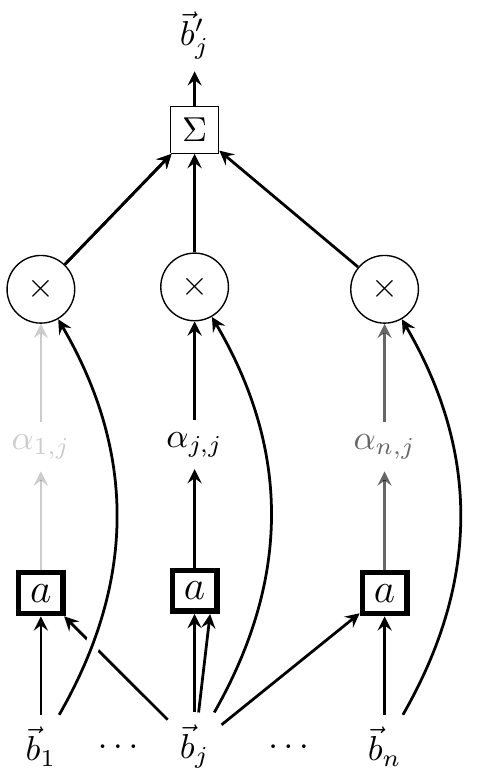}} \hfill
	\includegraphics[width=0.4\linewidth]{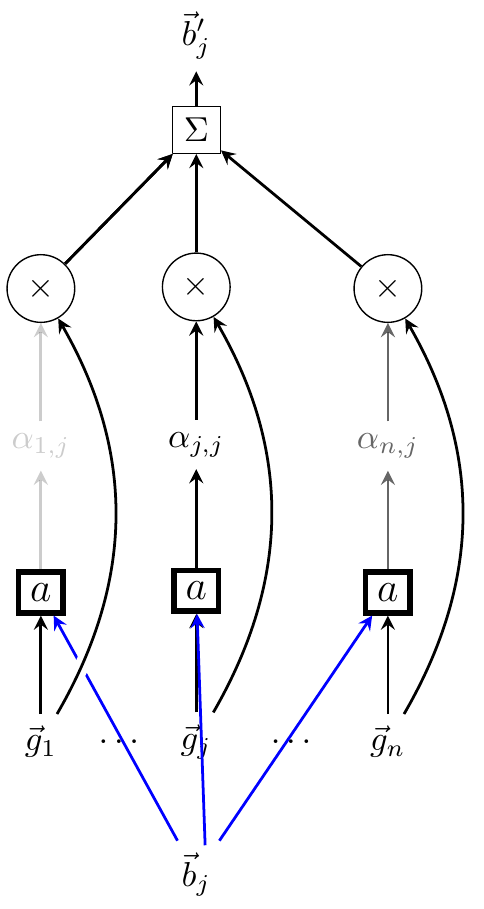}
	\label{attnmech}
	\caption{{\bf Left:} The \emph{self-attentional layer} of Fast-Parapred. {\bf Right:} The \emph{cross-modal attentional} layer of AG-Fast-Parapred.}
\end{figure}

The leveraged self-attention mechanism (depicted in Figure 2) is the same as the one utilised by \citet{GAT}. Taking a set of intermediate antibody residue features $ {\bf b} =\{ \vec{b}_{1}, \vec{b}_{2}, ... \vec{b}_{n} \}$, a shared neural network is applied to all pairs of residues, producing \emph{attention coefficients}:
\begin{equation}
	e_{ij} = a(\vec{b}_i, \vec{b}_j)
\end{equation}
indicating the importance of residue $j$'s features to residue $i$. Here the neural network $a$ is a single-layer feedforward neural network, parametrised by a weight vector $\vec{\bf a}$, and applying the LeakyReLU nonlinearity (with negative input slope $\alpha=0.2$):
\begin{equation}
	e_{ij} = \mathrm{LeakyReLU}\left(\vec{\bf a}^T[{\bf W}\vec{b}_i\|{\bf W}\vec{b}_j]\right)
\end{equation}
where $\cdot^T$ represents transposition and $\|$ is the concatenation operation. Here, ${\bf W}$ is a shared, learnable linear transformation of the residue features (preserving their dimensionality at 256)---adding further expressivity to the layer.

Once computed, the attention coefficients are \emph{normalised} using the softmax function, for easy comparability across different residues:
\begin{equation}
	\alpha_{ij} = \frac{\exp(e_{ij})}{\sum_{k=1}^n \exp(e_{ik})}
\end{equation}
Lastly, using the normalised attention coefficients, we compute a linear combination of all antibody residues' features for each attending antibody residue:
\begin{equation}
	\vec{b}'_{i} = \sigma \left(\sum_{j=1}^n \alpha_{ij} {\bf W}\vec{b}_{j}\right)
\end{equation}
which represents the final output of the layer  (summarised by Figure 3 (left)).

The regularisation methods used in this architecture are:
\begin{itemize}
	\item L$_2$-regularisation (with $\lambda=0.01$);
	\item Dropout \citep{Dropout} (with $p = 0.5$ on the final layer and $p = 0.15$ on all the other ones);
	\item Batch normalisation \citep{BatchNormalisation} on the output of each layer;
	\item A skip connection \citep{ResidualConnection} over self-attention, to preserve positional information of the residues.
\end{itemize}

The model (as well as all subsequent models) is trained using the Adam SGD optimiser---with base learning rate of 0.01 and other hyperparameters as presented in \citet{kingma2014adam}---for 20 epochs with a batch size of 32.

\begin{table*}[ht]
\centering
\caption{Comparative evaluation results with highlighted 95\% confidence intervals, after ten runs of 10-fold crossvalidation.}
\begin{tabular}{l c c c c}
\toprule 
 {\bf Method} & {\bf ROC AUC} & {\bf MCC} & {\bf Epoch time}\\ 
 \midrule
 {\bf proABC} \citep{proABC} & $0.851$ & $0.522$ & ---\\
{\bf Parapred} \citep{Parapred} & $0.880 \pm 0.002$ & $0.564 \pm 0.007$ & $0.190 \pm 0.019$s\\
{\bf Fast-Parapred} (ours) & $0.883 \pm 0.001$ & $0.572 \pm 0.004$ & ${\bf 0.085} \pm 0.015$s\\ 
{\bf AG-Fast-Parapred} (ours) &  ${\bf 0.899} \pm 0.004$ & ${\bf 0.598} \pm 0.012$ & $0.178 \pm 0.020$s\\
\bottomrule
\end{tabular}
\label{results}
\end{table*}

\subsection{Antibody-Antigen}

With similar motivation as before, we extract features from antibody and antigen amino acid residues by applying, independently to both, a stack of three \`{a} trous convolutional layers (with exactly the same hyperparameters as for the antibody-only model). The self-attention in the antibody-only paratope predictor is then replaced with \emph{cross-modal attention} of the antibody over the antigen residue features. We will refer to this architecture (presented in Figure 4) as \emph{AG-Fast-Parapred} for the remainder of this document.

\begin{figure}
	\includegraphics[width=\linewidth]{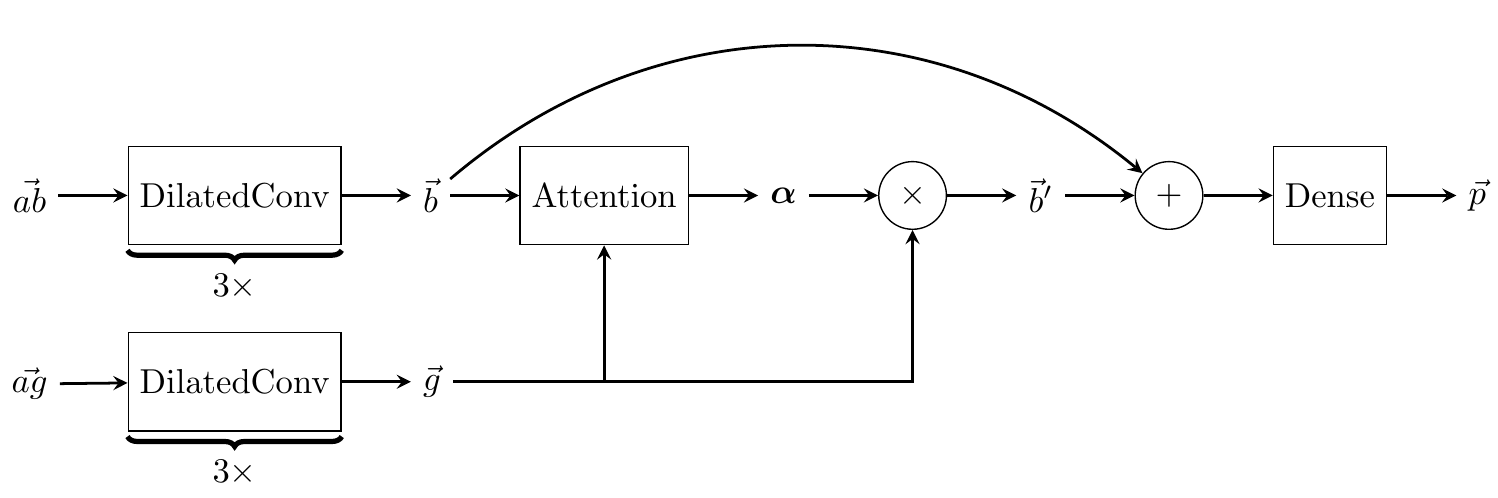}
	\caption{The AG-Fast-Parapred architecture.}
\end{figure}

We will focus on describing our \emph{cross-modal attentional layer} here, as the other layers are defined exactly the same as in Fast-Parapred (with identical hyperparameters). The input to the layer is a set of antibody residue features  
$ {\bf b} =\{ \vec{b}_{1}, \vec{b}_{2}, ... \vec{b}_{M} \} $, a set of antigen residue features $ {\bf g} =\{ \vec{g}_{1}, \vec{g}_{2}, ... \vec{g}_{N} \} $ and for each antibody residue $\vec{b}_{i} $ a set $\nu_{i}$ which marks the antigen residues which are in a fixed-range neighbourhood from $\vec{b}_{i} $. This neighbourhood was chosen to restrict the number of antigen residues being attended over by any antibody residue to 150. The attentional coefficients are then computed using the same attention mechanism $a$ as before, using the antibody residues as the \emph{queries} and antigen features as \emph{keys and values}. In addition, we now apply \emph{two} learned linear transformations (parametrised by ${\bf W}_1$ and ${\bf W}_2$) to each residue in ${\bf b}$ and ${\bf g}$, respectively. The attentional coefficients are subsequently normalised using a softmax activation, fully expanded out as follows:
\begin{equation}
 \alpha_{ij} = \frac{ \exp \left(\mathrm{LeakyReLU}\left(\vec{\bf a}^T[{\bf W}_1\vec{b}_i\|{\bf W}_2\vec{g}_j]\right)\right) } {\sum_{k \in \nu_{i} } \exp \left(\mathrm{LeakyReLU}\left(\vec{\bf a}^T[{\bf W}_1\vec{b}_i\|{\bf W}_2\vec{g}_k]\right)\right) }
\end{equation} 
Using the normalised attention coefficients, we then compute a linear combination of the corresponding antigen residues in the neighbourhood, for each attending antibody:
\begin{equation}
	\vec{b}'_{i} = \sigma \left({\sum_{j \in \nu_{i}}} \alpha_{ij} {\bf W}_2 \vec{g}_{j}\right)
\end{equation}
conveniently summarised by Figure 3 (right).

The result is, in a similar way to the Antibody-only method, passed through a pointwise convolutional layer and a logistic sigmoid non-linearity is applied, in order to classify each considered antibody amino acid residue as binding or non-binding.

We apply the same regularisation as for the antibody-only model---along with a skip connection over the cross-modal attention (which is in this case \emph{critical}, as the layer entirely discards antibody features).

\section{Results}
\label{results}

\subsection{Quantitative Results}

We perform ten runs of 10-fold crossvalidation (with 10 distinct splits of the data into 10 folds) on Parapred, Fast-Parapred and AG-Fast-Parapred. For each, we monitor ROC-AUC, Matthews correlation coefficient (which we also report for \emph{proABC}; Table 1), the wall-clock time it takes to perform one epoch of training, and the precision/recall curve (which we also report for \emph{Antibody i-Patch}; Figure 5), along with \emph{95\% confidence intervals}. Our results successfully demonstrate that:
\begin{itemize}
	\item Fast-Parapred has achieved the state-of-the-art-level result	 on antibody-only paratope prediction, while requiring only \emph{half} the computational time of Parapred;
	\item \emph{AG-Fast-Parapred} has significantly outperformed this result, for the first time successfully leveraging antigen information in a deep paratope predictor, while relying solely on convolutional and attentional layers, removing the dependency on recurrent layers entirely.
	\item It should be noted that AG-Fast-Parapred still improves on the epoch time of Parapred, despite working with input sizes that are up to $40\times$ larger.
\end{itemize}
\begin{figure}
\centering
\includegraphics[width=\linewidth]{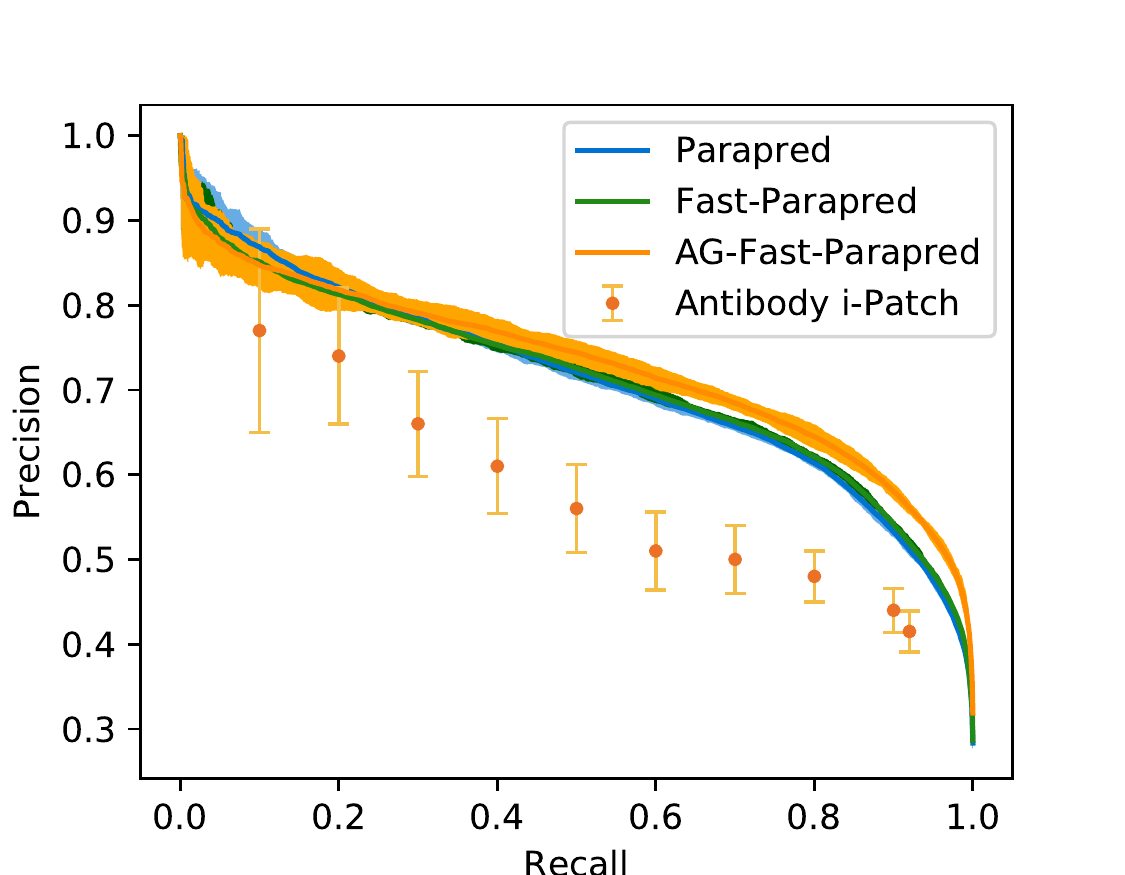}\label{labfig1}
\caption{Precision-Recall curves with highlighted 95\% confidence intervals, after ten runs of 10-fold crossvalidation.}
\label{fig:image}
\end{figure}

\subsection{Qualitative Results}

\begin{figure}
\centering
  \includegraphics[width=0.49\linewidth]{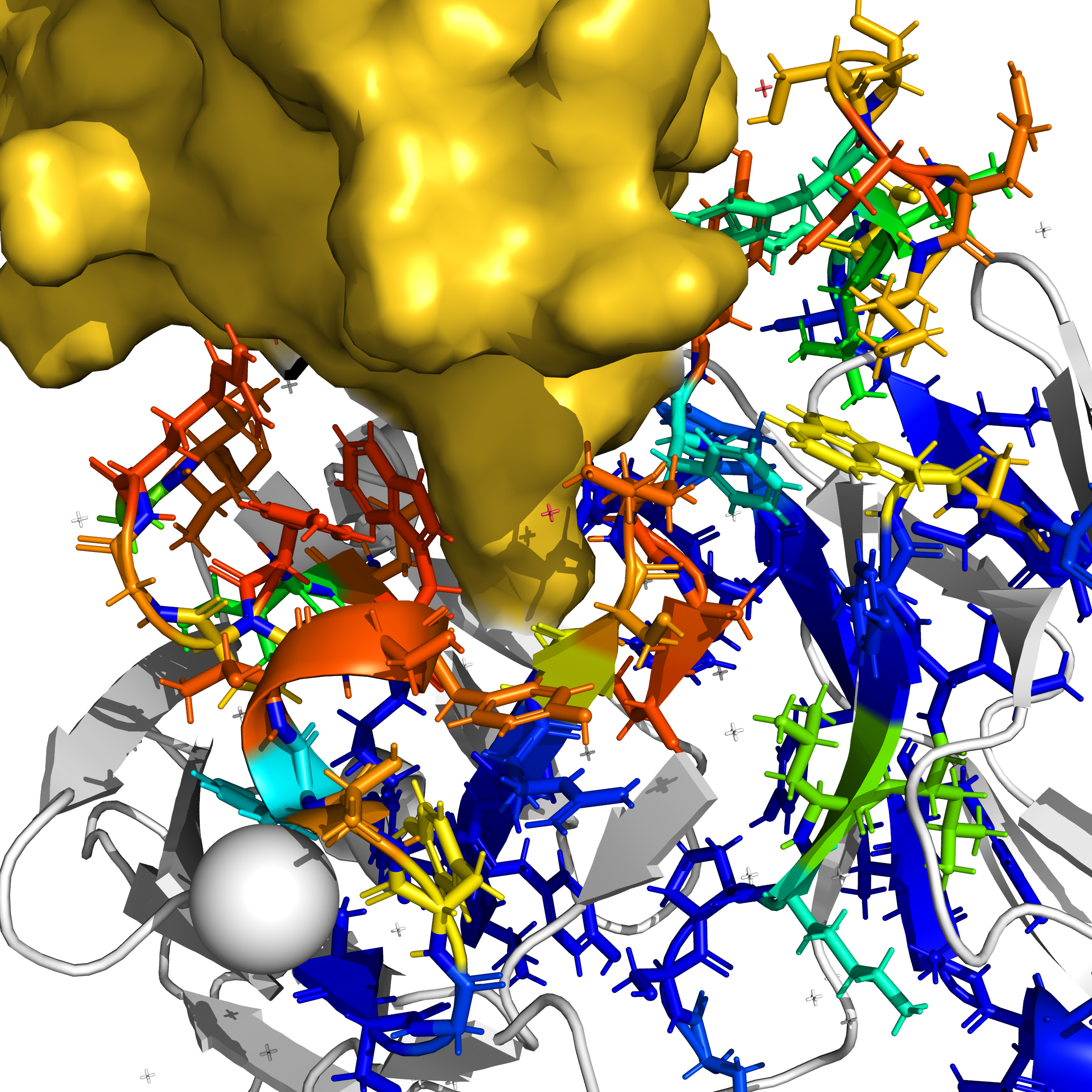} \hfill
  \includegraphics[width=0.49\linewidth]{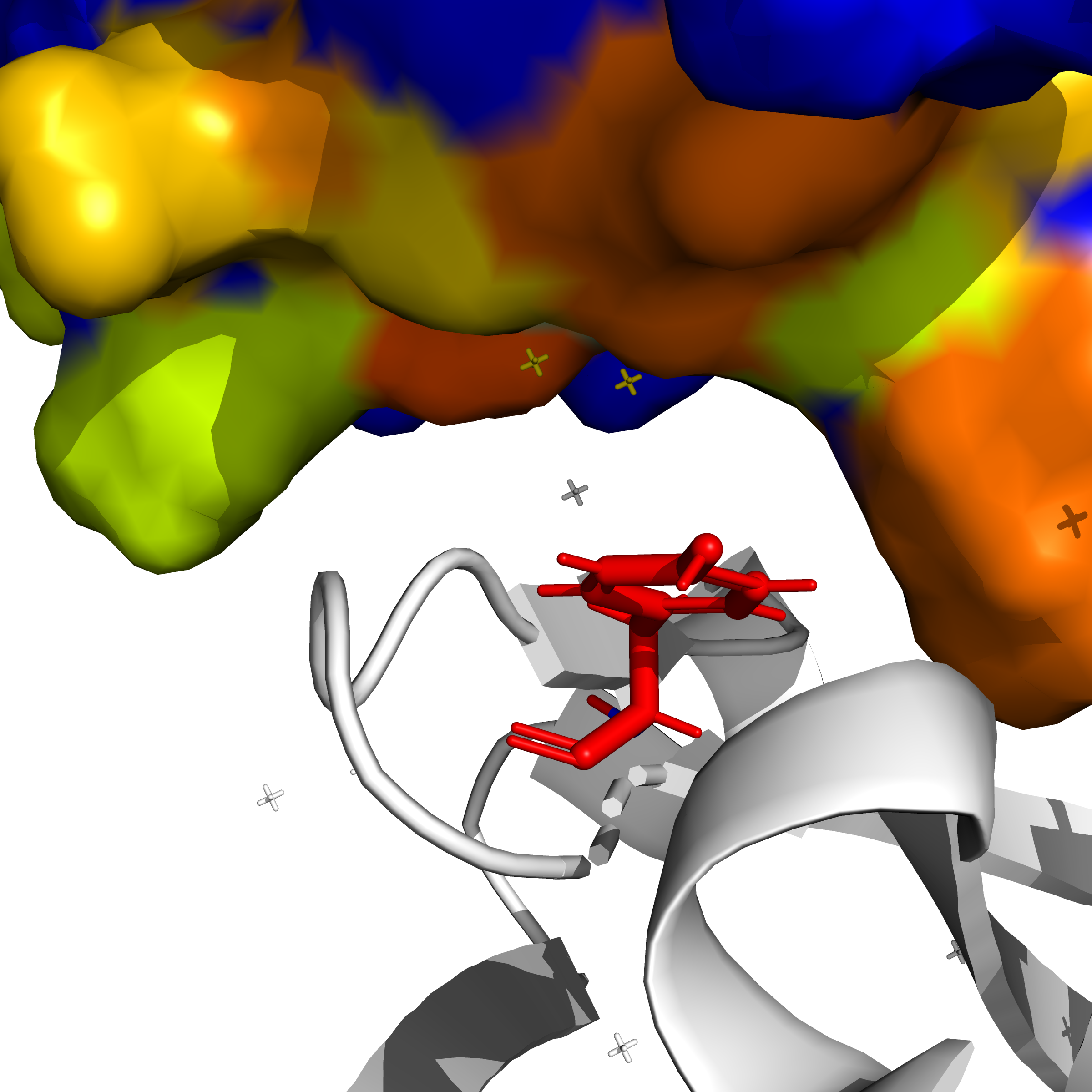}
  \caption{Best viewed in colour. For a test antibody-antigen complex: {\bf Left:} Antibody residue binding probabilities to the antigen (in gold) assigned by AG-Fast-Parapred. {\bf Right:} Normalised antigen attention weights for a single (binding) antibody residue (in red). Warmer colours indicate higher probabilities/coefficients.}
  \label{fig:test1}
\end{figure}

We visualise, using PyMOL, the computed binding probabilities of \emph{AG-Fast-Parapred}, on a test antibody-antigen complex, in Figure \ref{fig:test1} (left)---revealing that its neural network has learnt to appropriately infer positional information (predicting higher probabilities for the residues closer to the antigen), \emph{without being given any 3D coordinates}. The attentional coefficients computed by AG-Fast-Parapred are also visualised, for a single antibody residue, in Figure \ref{fig:test1} (right). From these we may observe that the attentional mechanism will tend to assign larger importances to antigen residues that are \emph{closer} to this antibody residue---indicating the usefulness of the cross-modal attentional mechanism, and potentially hinting at a joint method for predicting antigen binding sites (\emph{epitopes}), which we leave for future work.

\bibliography{example_paper}
\bibliographystyle{icml2018}

\end{document}